\documentclass{article}

\usepackage[preprint,nonatbib]{neurips_2019}

\usepackage[utf8]{inputenc} 
\usepackage[T1]{fontenc}    
\usepackage{booktabs}       
\usepackage{amsfonts}       
\usepackage{nicefrac}       
\usepackage{microtype}      

\usepackage[backend=bibtex,style=numeric-comp,maxbibnames=99]{biblatex}

\usepackage{amsmath,amssymb,amsfonts,amsthm} 
\usepackage[svgnames, table]{xcolor} 

\usepackage{graphicx}

\usepackage{float} 

\usepackage{bm} 
\usepackage{bbm, dsfont}

\newcommand{\mb}[1]{\mathbf{#1}}

\newcommand\blfootnote[1]{%
  \begingroup
  \renewcommand\thefootnote{}\footnote{#1}%
  \addtocounter{footnote}{-1}%
  \endgroup
}

\title{GMLS-Nets: A framework for learning from unstructured data}

\author{%
Nathaniel  Trask$^{1,+}$,   \hspace{1cm} Ravi G.~Patel$^{1}$, 
\hspace{1cm} 
Ben J.~Gross$^{2}$,  \hspace{1cm} Paul J.~Atzberger$^{2,\dagger}$
\\
\\
$^1$ Sandia National Laboratories \hspace{1cm} $^2$ University of California Santa Barbara \\ 
\hspace{-1.5cm}Center for Computing Research\hspace{1.9cm}$^\dagger$\texttt{atzberg@gmail.com \hspace{4.0cm}}\\
\hspace{-0.5cm}$^+$\texttt{natrask@sandia.gov}
\hspace{2.2cm} \url{http://atzberger.org/} \\
\\
September 5, 2019
}

\begin{document}

\maketitle

\begin{abstract}
Data fields sampled on irregularly spaced points arise in many applications in the sciences and engineering.  For regular grids, Convolutional Neural Networks (CNNs) have been successfully used to gaining benefits from weight sharing and invariances.  We generalize CNNs by introducing methods for data on unstructured point clouds based on Generalized Moving Least Squares (GMLS).  GMLS is a non-parametric technique for estimating linear bounded functionals from scattered data, and has recently been used in the literature for solving partial differential equations.  By parameterizing the GMLS estimator, we obtain learning methods for operators with unstructured stencils.  In GMLS-Nets the necessary calculations are local, readily parallelizable, and the estimator is supported by a rigorous approximation theory.  We show how the framework may be used for unstructured physical data sets to perform functional regression to identify associated differential operators and to regress quantities of interest.  The results suggest the architectures to be an attractive foundation for data-driven model development in scientific machine learning applications. 
\end{abstract}


\section{Introduction}
Many scientific and engineering applications require processing data sets sampled on irregularly spaced points.  Consider e.g. GIS data associating geospatial locations with measurements, LIDAR data characterizing object geometry via point clouds, scientific simulations with unstructured meshes.  This need is amplified by the recent surge of interest in scientific machine learning (SciML) \cite{DOEReportAI2018} targeting the application of data-driven techniques to the sciences. In this setting, data typically takes the form of e.g. synthetic simulation data from meshes, or from sensors associated with data sites evolving under unknown or partially known dynamics. This data is often scarce or highly constrained, and it has been proposed that successful SciML strategies will leverage prior knowledge to enhance information gained from such data~\cite{Atzberger_PosPaper_2018,DOEReportAI2018}. One may exploit physical properties and invariances such as transformation symmetries, conservation structure, or mathematical knowledge such as solution regularity~\cite{BrennerDataDrivenPDE2019,BruntonKutz2016,Atzberger_PosPaper_2018}. This new application space necessitates ML architectures capable of utilizing such knowledge.

\blfootnote{Implementations in TensorFlow and PyTorch are available at \url{https://github.com/rgp62/gmls-nets} and \url{https://github.com/atzberg/gmls-nets}.\\
1. Sandia National Laboratories is a multimission laboratory managed and operated by National Technology and Engineering Solutions of Sandia, LLC.,a wholly owned subsidiary of Honeywell International, Inc., for the U.S. Department of Energy’s National Nuclear Security Administration under contract DE-NA-0003525. This paper describes objective technical results and analysis. Any subjective views or opinions that might be expressed in the paper do not necessarily represent the views of the U.S. Department of Energy or the United States Government. \\
* Work supported by DOE Grant ASCR PhILMs $\mbox{DE-SC0019246.}$}

For data sampled on regular grids, Convolutional Neural Networks (CNNs) are widely used to exploit translation invariance and hierarchical structure to extract features from data. Here we generalize this technique to the SciML setting by introducing \textit{GMLS-Nets} based on the scattered data approximation theory underlying generalized moving least squares (GMLS). Similar to how CNNs learn stencils which benefit from weight-sharing, GMLS-Nets operate by using local reconstructions to learn operators between function spaces. The resulting architecture is similarly interpretable and serves as an effective generalization of CNNs to unstructured data, while providing mechanisms to incorporate knowledge of underlying physics.

In this work we show how GMLS-Nets may be used in a SciML setting. Our results show GMLS-Nets are an effective tool to discover partial diferential equations (PDEs), which may be used as a foundation to construct data-driven models while preserving physical invariants like conservation principles. We also show they may be used to improve traditional scientific components, such as time integrators. We show they also can be used to regress engineering quantities of interest from scientific simulation data.  Finally, we briefly show GMLS-Nets can perform reasonably relative to convNets on traditional computer vision benchmarks.  These results indicate the promise of GMLS-Nets to support data-driven modeling efforts in SciML applications.  Implementations in TensorFlow and PyTorch are available at \url{https://github.com/rgp62/gmls-nets} and \url{https://github.com/atzberg/gmls-nets}.

\subsection{Generalized Moving Least Squares (GMLS)}
Generalized Moving Least Squares (GMLS) is a non-parametric functional regression technique to construct approximations of linear, bounded functionals from scattered samples of an underlying field by solving local least-square problems.  On a Banach space $\mathbb{V}$ with dual space $\mathbb{V}^*$, we aim to recover an estimate of a given target functional $\tau_{\tilde{\mb{x}}}[u]\in\mathbb{V}^*$ acting on $u=u(\mb{x})\in\mathbb{V}$,  where $\mb{x},\tilde{\mb{x}}$ denote associated locations in a compactly supported domain $\Omega\subset\mathbb{R}^d$. We assume $u$ is characterized by an unstructured collection of sampling functionals, $\Lambda(u) := \left\{\lambda_j(u)\right\}_{j=1}^N \subset \mathbb{V}^*$. 

To construct this estimate, we consider $\mathbb{P}\subset \mathbb{V}$ and seek an element $p^* \in \mathbb{P}$ which provides an optimal reconstruction of the samples in the following weighted-$\ell_2$ sense.
\begin{equation}
\label{eqn:opt_gmls_reconstruct}
    p^* = \underset{{p \in \mathbb{P}}}{\mbox{argmin}} \sum_{j=1}^N \left( \lambda_j(u) -\lambda_j(p) \right)^2 \omega(\lambda_j,\tau_{\tilde{\mb{x}}}).
\end{equation}
Here $\omega(\lambda_j,\tau_{\tilde{\mb{x}}})$ is a positive, compactly supported kernel function establishing spatial correlation between the target functional and sampling set. If one associates locations $\mathbb{X}_h := \left\{\mathbf{x}_j\right\}_{j=1}^N \subset{\Omega}$ with $\Lambda(u)$, then one may consider radial kernels $\omega = W_\epsilon(||\mb{x}_j - \tilde{\mb{x}}||_2)$, with support $r < \epsilon$.

Assuming the basis $\mathbb{P} = \mbox{span}\{\phi_1,...,\phi_{\dim(\mathbb{P})} \}$, and denoting $\Phi(x)=\left\{\phi_i(x)\right\}_{i=1,...,dim(\mathbb{P})}$, the optimal reconstruction may be written in terms of an optimal coefficient vector $\mathbf{a}(u)$
\begin{equation}
\label{eqn:gmls_coeff_a}
    p^* = \Phi(x)^\intercal \mathbf{a}(u).
\end{equation}

Provided one has knowledge of how the target functional acts on $\mathbb{P}$, the final GMLS estimate may be obtained by applying the target functional to the optimal reconstruction
\begin{equation}
\label{eqn:gmls_approx_tau1}
\tau^h_{\tilde{\mb{x}}}[u] = \tau_{\tilde{\mb{x}}}(\Phi)^\intercal \mathbf{a}(u).
\end{equation}

Sufficient conditions for the existence of solutions to Eqn. \ref{eqn:opt_gmls_reconstruct} depend only upon the unisolvency of $\Lambda$ over $\mathbb{V}$, the distribution of samples $\mathbb{X}_h$, and mild conditions on the domain $\Omega$; they are independent of the choice of $\tau_{\tilde{\mb{x}}}$. For theoretical underpinnings and recent applications, we refer readers to~\cite{wendland2004scattered,trask2017high,trask2019conservative,Atzberger_GMLS_Manifold_2019}.

GMLS has primarily been used to obtain point estimates of differential operators to develop meshfree discretizations of PDEs. The abstraction of GMLS however provides a mathematically rigorous approximation theory framework which may be applied to a wealth of problems, 
whereby one may tailor the choice of $\tau_{\tilde{\mb{x}}}$, $\Lambda$, $\mathbb{P}$ and $\omega$ to a given application. In the current work, we will assume the action of $\tau_{\tilde{\mb{x}}}$ on $\mathbb{P}$ is unknown, and introduce a parameterization $\tau_{\tilde{\mb{x}},\xi}(\Phi)$, where $\xi$ denote hyperparameters to be inferred from data. Classically, GMLS is restricted to linear bounded target functionals; we will also consider a novel nonlinear extension by considering estimates of the form
\begin{equation}
\label{eqn:gmls_approx_tau1_nonlinear}
\tau^h_{\tilde{\mb{x}}}[u] = q_{\tilde{\mb{x}},\xi}(\mathbf{a}(u)),
\end{equation}
where $q_{\tilde{\mb{x}},\xi}$ is a family of nonlinear operators parameterized by $\xi$ acting upon the GMLS reconstruction. Where unambiguous, we will drop the $\tilde{\mb{x}}$ dependence of operators and simply write e.g.  $\tau^h[u] = q_{\xi}(\mathbf{a}(u))$.  We have recently used related non-linear variants of GMLS to develop solvers for PDEs on manifolds in~\cite{Atzberger_GMLS_Manifold_2019}.

For simplicity, in this work we specialize as follows. Let: $\Lambda$ be point evaluations on $\mathbb{X}_h$; $\mathbb{P}$ be $\pi_m(\mathbb{R}^d)$, the space of $m^{th}$-order polynomials; let $W_\epsilon(r) = \left(1 - {r}/{\epsilon}\right)^{\bar{p}}_+$, where $f_+$ denotes the positive part of a function $f$ and $p \in \mathbb{N}$. We stress however that this framework supports a much broader application. Consider e.g. learning from flux data related to $H(div)$-conforming discretizations, where one may select as sampling functional $\lambda_i(\mathbf{u}) = \int_{f_i} \mathbf{u} \cdot d\mathbf{A}$, or consider the physical constraints that may be imposed by selecting $\mathbb{P}$ as be divergence free or satisfy a differential equation. 

We illustrate now the connection between GMLS and convolutional networks in the case of a uniform grid, $\mathbb{X}_h \subset \mathbb{Z}^d$. Consider a sampling functional $\lambda(u) = \left(u(\mathbf{x}_j) - u(\mathbf{x}_i)\right)$, and assume the parameterization $\tau_{\tilde{\mb{x}},\xi}(\Phi)=\left<\xi_1,...,\xi_{dim(\mathbb{P})}\right>$, $\mb{x}_{i,j} = \mb{x}_{i} - \mb{x}_{j}$. Then the GMLS estimate is given explicitly at a point $\mb{x}_i$ by
\begin{equation}
    \begin{split}
        \tau^h_{\tilde{\mb{x}_i}}[u] = \sum_{\alpha,\beta,j} \xi_\alpha \left(\sum_k \phi_\alpha (\mb{x}_k) W(\mb{x}_{ik})\phi_\beta (\mb{x}_k)\right)^{-1} \\
        \phi_\beta (\mb{x}_j) W(\mb{x}_{i,j}) (u_j-u_i).
    \end{split}
\end{equation}

Contracting terms involving $\alpha,\beta$ and $k$, we may write $\tau^h_{\tilde{\mb{x}_i}}[u] = \sum_{j} c(\tau,\Lambda)_{ij} (u_j-u_i)$. The collection of stencil coefficients at $x_i \in \mathbb{X}_h$ are $\left\{c(\tau,\Lambda)_{ij}\right\}_j$. Therefore, one application for GMLS is to build stencils similar to convolutional networks. A major distinction is that GMLS can handle scattered data sets and a judicious selection of $\Lambda$, $\mathbb{P}$ and $\omega$ can be used to inject prior information. Alternatively, one may interpret the regression over $\mathbb{P}$ as an encoding in a low-dimensional space well-suited to characterize common operators. For continuous functions for example, an operator's action on the space of polynomials is often sufficient to obtain a good approximation. We also remark that unlike CNNs there is often less need to handle boundary effects; GMLS-nets is capable of learning one-sided stencils.

\subsection{GMLS-Nets}
\begin{figure}[h]
\centering
\includegraphics[width=0.65\columnwidth]{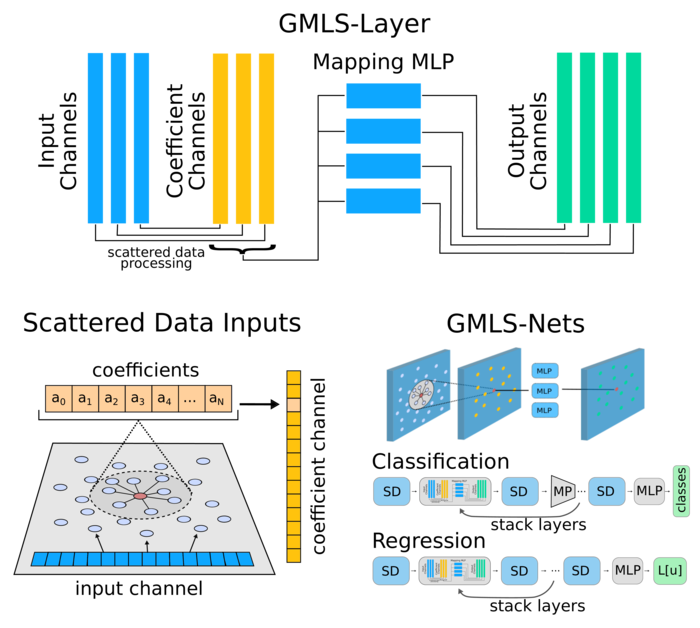}
\caption{GMLS-Nets.  Scattered data inputs are processed by learnable operators $\tau[u]$ parameterized via GMLS estimators.  A local reconstruction is built about each data point and encoded as a coefficient vector via equation~\ref{eqn:gmls_coeff_a}.  The coefficient mapping $q(\mb{a})$ of equation~\ref{eqn:gmls_approx_tau1_nonlinear} provides the learnable action of the operator.  GMLS-Layers can be stacked to obtain deeper architectures and combined with other neural network operations to perform classification and regression tasks \textit{(inset, SD: scattered data, MP: max-pool, MLP: multi-layer perceptron)}.
}
\label{fig:gmls_layers}  
\end{figure}

From an ML perspective, GMLS estimation consists of two parts: (i) data is encoded via the coefficient vector $\mathbf{a}(u)$ providing a compression of the data in terms of $\mathbb{P}$, (ii) the operator is regressed over $\mathbb{P}^*$; this is equivalent to finding a function $\mathbf{q}_\xi:\mathbf{a}(u)\rightarrow\mathbb{R}$.  We propose GMLS-Layers encoding this process in Figure~\ref{fig:gmls_layers}. 

This architecture accepts input channels indexed by $\alpha$ which consist of components of the data vector-field $[\mb{u}]^{\alpha}$ sampled over the scattered points $\mathbb{X}_h$.  We allow for different sampling points for each channel, which may be helpful for heterogeneous data.  Each of these input channels is then used to obtain an encoding of the input field as the vector $\mathbf{a}(u)$ identifying the optimal representer in $\mathbb{P}$. 

We next select our parameterization of the functional via $q_\xi$, which may be any family of functions trainable by back-propagation. We will consider two cases in this work appropriate for linear and non-linear operators. In the linear case we consider $\bm{q}_xi(\mb{a}) = \mathbf{\xi}^T\mb{a}$, which is sufficient to exactly reproduce differential operators. For the nonlinear case we parameterize with a multi-layer perceptron (MLP), $\bm{q}_
\xi(\mb{a}) = \mbox{MLP}(\mb{a})$. Note that in the case of linear activation function, the single layer MLP model reduces to the linear model.

Nonlinearity may thus be handled within a single nonlinear GMLS-Layer, or by stacking multiple linear GMLS-layers with intermediate ReLU's, the later mapping more directly onto traditional CNN construction. We next introduce pooling operators applicable to unstructured data, whereby for each point in a given target point cloud $\mathbb{X}_h^{target}$, $\phi(x_i) = F\left(\{x_j | {j \in \mathbb{X}_h},{|x_j-x_i| < \epsilon}  \}\right)$.  Here $F$ represents the pooling operator (e.g. max, average, etc.). With this collection of operators, one may construct architectures similar to CNNs by stacking GMLS-Layers together with pooling layers and other NN components. Strided GMLS-layers generalizing strided CNN stencils may be constructed by choosing target sites on a second, smaller point cloud.

\subsection{Relation to other work.}
Many recent works aim to generalize CNNs away from the limitations of data on regular grids~\cite{SpectralCNNs_Bruna_Lecun_2014,Bronstein_BeyondEuc_2017}.  This includes work on handling inputs in the form of directed and un-directed graphs~\cite{GraphNNs_Scarselli_2009}, processing graphical data sets in the form of meshes and point-clouds~\cite{Guibas_PointNet_PlusPlus_2017,DeepSets2017}, and in handling scattered sub-samplings of images~\cite{SplineCNN_Fey_2018,SpectralCNNs_Bruna_Lecun_2014}.  Broadly, these works: (i) use the spectral theory of graphs and generalize convolution in the frequency domain~\cite{SpectralCNNs_Bruna_Lecun_2014}, (ii) develop localized notions similar to convolution operations and kernels in the spatial domain~\cite{KPConv_Hugues_2019}.  GMLS-Nets is most closely related to the second approach.

The closest works include SplineCNNs~\cite{SplineCNN_Fey_2018}, MoNet~\cite{Monti_GeometricDL_2016, KipfWelling_Kernel_CNNs_2016}, KP-Conv~\cite{KPConv_Hugues_2019}, and SpiderCNN~\cite{SpiderCNN_Xu_2018}.
In each of these methods a local spatial convolution kernel is approximated by a parameterized family of functions: open/closed B-Splines~\cite{SplineCNN_Fey_2018}, a Gaussian correlation kernel~\cite{Monti_GeometricDL_2016,KipfWelling_Kernel_CNNs_2016}, or a kernel function based on a learnable combination of radial ReLu's~\cite{KPConv_Hugues_2019}.  The SpiderCNNs share many similarities with GMLS-Nets using a kernel that is based on a learnable degree-three Taylor polynomial that is taken in product with a learnable radial piecewise-constant weight function~\cite{SpiderCNN_Xu_2018}.  
A key distinction of GMLS-Nets is that operators are regressed directly over the dual space $\mathbb{V}^*$ without constructing shape/kernel functions. Both approaches provide ways to approximate the action of a processing operator that aggregates over scattered data.

We also mention other meshfree learning frameworks: PointNet~\cite{Guibas_PointNet_2017,Guibas_PointNet_PlusPlus_2017} and Deep Sets~\cite{DeepSets2017}, but these are aimed primarily at set-based data and geometric processing tasks for segmentation and classification. Additionally, Radial Basis Function (RBF) networks are similarly built upon similar approximation theory ~\cite{RBF_Early_Broomhead_1988,RBF_NNs_Poggio_1990}. 

Related work on operator regression in a SciML context include~\cite{Karniadakis_PINNs_2019,
Karniadakis_Raissi_HiddenPhys_2018,PDENet_Long2018,BruntonKutz2016,
RudyKutz2017,Lagaris_PDE_ODE_NN_1998,
BrennerDataDrivenPDE2019,Patel2018}.  In PINNs~\cite{Karniadakis_PINNs_2019,Karniadakis_Raissi_HiddenPhys_2018}, a versatile framework based on DNNs is developed to regress both linear and non-linear PDE models while exploiting physics knowledge. In~\cite{BrennerDataDrivenPDE2019} and PDE-Nets~\cite{PDENet_Long2018}, CNNs are used to learn stencils to estimate operators.  In~\cite{BruntonKutz2016,RudyKutz2017} dictionary learning is used along with sparse optimization methods to identify dynamical systems to infer physical laws associated with time-series data. In~\cite{Patel2018}, regression is performed over a class of nonlinear pseudodifferential operators, formed by composing neural network parameterized Fourier multipliers and pointwise functionals.

GMLS-Nets can be used in conjunction with the above methods. GMLS-Nets have the distinction of being able to move beyond reliance on CNNs on regular grids, no longer need moment conditions to impose accuracy and interpretability of filters for estimating differential operators~\cite{PDENet_Long2018}, and do not require as 
strong assumptions about the particular form of the PDE or a pre-defined dictionary as in~\cite{Karniadakis_PINNs_2019,RudyKutz2017}. We expect that prior knowledge exploited globally in PINNs methods may be incorporated into the GMLS-Layers. In particular, the ability to regress natively over solver degrees of freedom will be particularly useful for SciML applications.

\section{Results}

\subsection{Learning differential operators and identifying governing equations.} 

\begin{figure}[h]
\centering
\includegraphics[width=0.45\columnwidth]{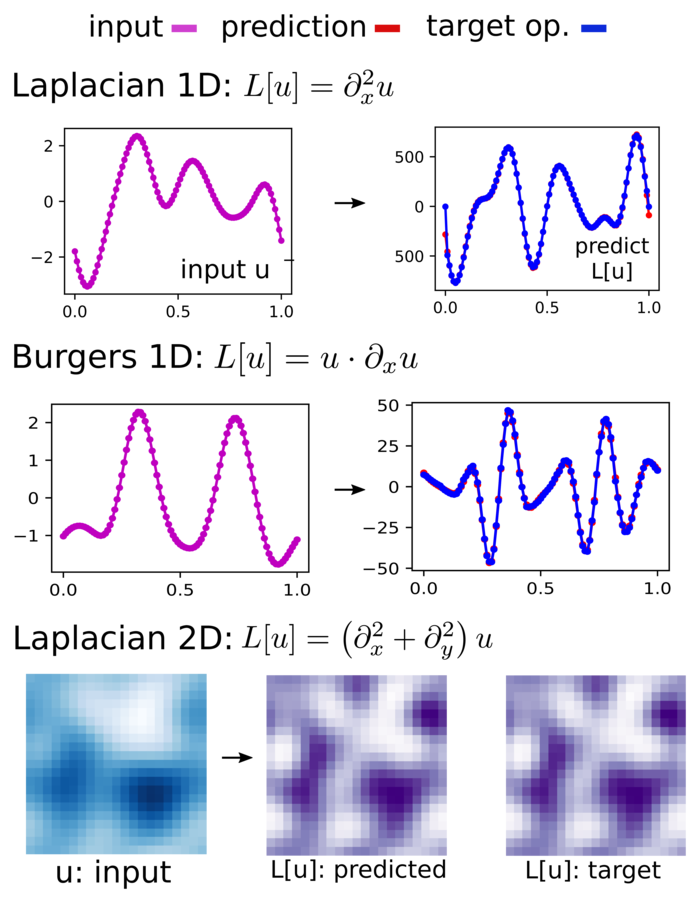}
\caption{Regression of Differential Operators.  GMLS-Nets can accurately learn both linear and non-linear operators, shown is the case of the 1D/2D Laplacians and Burger's equation.  In-homogeneous operators can also be learned by including as one of the input channels the location $x$.  Training and test data consists of random input functions in 1d at $10^2$ nodes on $[0,1]$ and in 2d at $400$ nodes in $[0,1]\times[0,1]$. Each random input function follows a Gaussian distribution with $u(\mb{x}) = \sum_{\mb{k}} \xi_k \exp\left(i2\pi\mb{k}\cdot \mb{x}/L\right)$ with $\xi_k \sim \exp(-\alpha_1k^2)\eta(0,1)$.  Training and test data is generated with $\alpha_1 = 0.1$ by computed operators with spectral accuracy for $N_{train} = 5\times10^4$ and $N_{test} = 10^4$.} 
\label{diff_op_reg}
\end{figure}

Many data sets arising in the sciences are generated by processes for which there are expected governing laws expressible in terms of ordinary or partial differential equations.  GMLS-Nets provide natural features to regress such operators from observed state trajectories or responses to fluctuations.  We consider the two settings
\begin{eqnarray}
\frac{\partial{u}}{\partial t} = \mathcal{L}[u(t,x)] \hspace{0.1cm} \mbox{and} \hspace{0.1cm}
\mathcal{L}[u(x)] = -f(x).
\end{eqnarray}
The $\mathcal{L}[u]$ can be a linear or non-linear operator.  When the data are snapshots of the system state $u^{n} = u(t^n)$ at discrete times $t^n = n\Delta{t}$, we use estimators based on
\begin{eqnarray}
\frac{u^{n+1} - u^{n}}{\Delta{t}} = \mathcal{L}[\{u^{k}\}_{k\in\mathcal{K}};\xi].
\label{equ_estimators}
\end{eqnarray}
In the case that $\mathcal{K} = \{n+1\}$, this corresponds to using an Implicit Euler scheme to model the dynamics. Many other choices are possible, and later we shall discuss estimators with conservation properties.  The learning capabilities of GMLS-Nets to regress differential operators are shown in Fig.~\ref{diff_op_reg}. As we shall discuss in more detail, this can be used to identify the underlying dynamics and obtain governing equations.

\subsection{Long-time integrators: discretization for native data-driven modeling.}

\begin{figure}[h]
    \centering
    \includegraphics[width=0.6\linewidth]{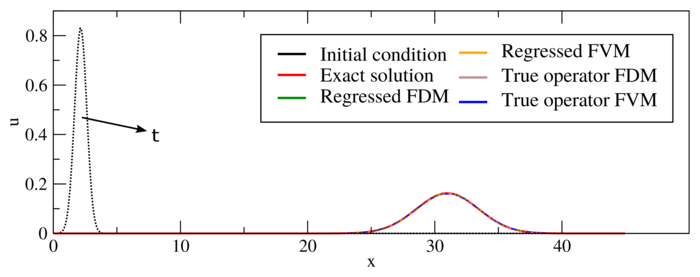}\\
    \includegraphics[width=0.6\linewidth]{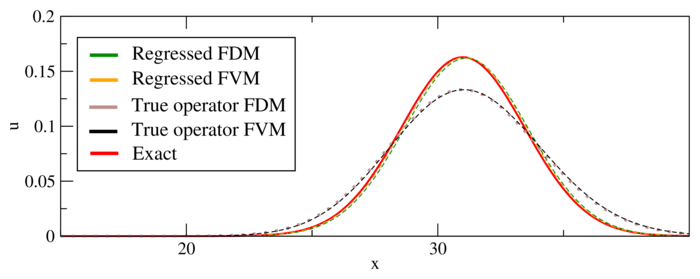}
    \caption{\textit{Top:} Advection-diffusion solution when $\Delta t = \Delta t_{CFL}$. The true model solution and regressed solution all agree with the analytic solution. \textit{Bottom:} Solution for under-resolved dynamics with $\Delta t = 10 \Delta t_{CFL}$. The implicit integrator causes FDM/FVM of true operator to be overly dissipative. The regressed operator matches well with the FVM operator, matching the phase almost exactly.}
    \label{fig:advdif}
\end{figure}
\begin{table}[h]
\centering
\begin{tabular}{ccccc}
\textbf{$\Delta t$/$\Delta t_{CFL}$} & \textbf{$\mathcal{L}_{FDM,ex}$} & \textbf{$\mathcal{L}_{FDM}$} & \textbf{$\mathcal{L}_{FVM,ex}$} & \textbf{$\mathcal{L}_{FVM}$} \\ \hline \\
0.1                      & 0.00093 & 0.00015     & 0.00014 & 0.00010     \\
1                        & 0.0011  & 0.00093     & 0.0011  & 0.00011     \\
10                       & 0.0083  & 0.0014      & 0.0083  & 0.00035  
\end{tabular}
\caption{The $\ell_2$-error for data-driven finite difference model (FDM) and finite volume models (FVM) for advection-diffusion equation. Comparisons made to classical discretizations using exact operators.  For conservative data-driven finite volume model, there is an order of magnitude better accuracy for large timestep integration.}
\label{advdiffL2err}
\end{table}

The GMLS framework provides useful ways to target and sample arbitrary functionals. In a data transfer context, this has been leveraged to couple heterogeneous codes. For example, one may sample the flux degrees of freedom of a Raviart-Thomas finite element space and target cell integral degrees of freedom of a finite volume code to perform \textit{native data transfer}.  This avoids the need to perform intermediate projections/interpolations \cite{kuberry2018virtual}. Motivated by this, we demonstrate that GMLS may be used to learn \textit{discretization native data-driven models}, whereby dynamics are learned in the natural degrees of freedom for a given model.  This provides access to structure preserving properties such as conservation, e.g., conservation of mass in a physical system.

We take as a source of training data the following analytic solution to the 1D unsteady advection-diffusion equation with advection and diffusion coefficients $a$ and $\nu$ on the interval $\Omega = [0,30]$.
\begin{equation}
    u_{ex}(x,t) = \frac{1}{a\sqrt{4 \pi \nu t}} \exp{\left(-\frac{x - (x_0 + a t)}{4 \nu t}\right)}
    \label{exactAD}
\end{equation}
To construct a finite difference model (FDM), we assume a node set $\mathbb{N} = \left\{x_0=0,x_1,...,x_{N-1},x_{N}=30\right\}$. To construct a finite volume model (FVM), we construct the set of cells $\mathbb{C} = \left\{ [x_i,x_{i+1}] , x_i,x_{i+1} \in \mathbb{N}, i \in \left\{0,...,N-1\right\}\right\}$, with associated cell measure $\mu(c_i) = |x_{i+1} - x_i|$ and set of oriented boundary faces $F_i = \partial c_i = \left\{x_{i+1},-x_i\right\}$. We then assume for uniform timestep $\Delta t = t^{n+1}-t^n$ the Implicit Euler update for the FDM given by
\begin{equation}
    \frac{u^{n+1}_i - u^n_i}{\Delta t} = \mathcal{L}_{FDM}[u^{n+1};\xi],
        \label{equ:estimator_fdm}
\end{equation}
To obtain conservation we use the FVM update
\begin{equation}
    \frac{u^{n+1}_i - u^n_i}{\Delta t} = \frac{1}{\mu(c_i)} \sum_{f\in F_i} \int \mathcal{L}_{FVM}[u^{n+1};\xi] \cdot d\mathbf{A}.
    \label{equ:estimator_fvm}
\end{equation}
For the advection-diffusion equation in the limit $\Delta t\rightarrow 0$, $\mathcal{L}_{FDM,ex} = a \cdot \nabla u + \nu \nabla^2 u$ and $\mathcal{L}_{FVM,ex} = a u + \nu \nabla u$. By construction, for any choice of hyperparameters $\xi$ the FVM will be locally conservative. In this sense, the physics of mass conservation are enforced strongly via the discretization, and we parameterize only an empirical closure for fluxes - GMLS naturally enables such native flux regression.

We use a single linear GMLS-net layer to parameterize both $\mathcal{L}_{FDM}$ and $\mathcal{L}_{FVM}$, and train over a single timestep by using Eqn. \ref{exactAD} to evaluate the exact time increment in Eqns. \ref{equ:estimator_fdm}-\ref{equ:estimator_fvm} .
We perform gradient descent to minimize the RMS of the residual with respect to $\xi$. For the FDM and FVM we use a cubic and quartic polynomial space, respectively. Recall that to resolve the diffusion and advective timescales one would select a timestep of roughly $\Delta t_{CFL} = \min\left(\frac12 \frac{a \Delta t}{\Delta x}, \frac14 \frac{\nu \Delta t}{\Delta x^2} \right)$. 

After regressing the operator, we solve the extracted scheme to advance from $\left\{u^0_i = u(x_i,t_0)\right\}_i$ to $\left\{ u^{t_{final}}_i \right\}_i$. As implicit Euler is unconditionally stable, one may select $\Delta t \gg \Delta t_{CFL}$ at the expense of introducing numerical dissipation, "smearing" the solution. We consider $\Delta t \in \left\{0.1 \Delta t_{CFL}, \Delta t_{CFL}, 10\Delta t_{CFL}\right\}$ and compare both the learned FDM/FVM dynamics to those obtained with a standard discretization (i.e. letting $\mathcal{L}_{FDM} = \mathcal{L}_{FDM,ex}$. From Fig. \ref{fig:advdif} we observe that for $\Delta t/\Delta t_{CFL} \leq 1$ both the regressed and reference models agree well with the  analytic solution. However, for $\Delta t = 10\Delta t_{CFL}$, we see that while the reference models are overly dissipative, the regressed models match the analytic solution. Inspection of the $\ell_2-$norm of the solutions at $t^{final}$ in Table \ref{advdiffL2err} indicates that as expected, the classical solutions corresponding to $\mathcal{L}_{FDM,ex}$ and $\mathcal{L}_{FVM,ex}$ converge as $O(\Delta t)$. The regressed FDM is consistently more accurate than the exact operator. Most interesting, the regressed FVM is roughly independent of $\Delta t$, providing a $20\times$ improvement in accuracy over the classical model. This preliminary result suggests that GMLS-Nets offer promise as a tool to develop non-dissipative implicit data-driven models. We suggest that this is due to the ability for GMLS-Nets to regress higher-order differential operator corrections to the discrete time dynamics, similar to e.g. Lax-Friedrichs/Lax-Wendroff schemes.

\subsection{Data-driven modeling from molecular dynamics.}
\begin{figure}[h]
\centering
\includegraphics[width=0.6\columnwidth]{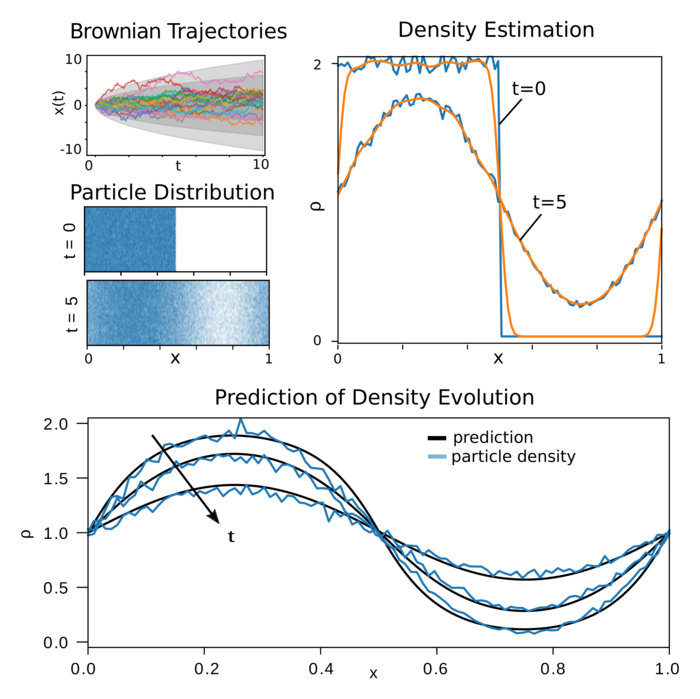}
\caption{GMLS-Nets can be trained with molecular-level data to infer continuum dynamical models.  Data are simulations of Brownian motion with periodic boundary conditions on $\Omega = [0,1]$ and diffusivity $D = 1$ \textit{(top-left, unconstrained trajectory)}.  Starting with initial density of a heaviside function, we construct histograms over time to estimate the particle density \textit{(upper-right, solid lines)} and perform further filtering to remove sampling noise \textit{(upper-right, dashed lines)}.  GMLS-Net is trained using FVM estimator of equation~\ref{equ:estimator_fvm}. Predictive continuum model is obtained for the density evolution.  Long-term agreement is found between the particle-level simulation \textit{(bottom, solid lines)} and the inferred continuum model \textit{(bottom, dashed lines)}.}
\label{brownian}
\end{figure}

In science and engineering applications, there are often high-fidelity descriptions of the physics based on molecular dynamics.  One would like to extract continuum descriptions to allow for predictions over longer time/length-scales or reduce computational costs.  Coarse-grained modeling efforts also have similar aims while retaining molecular degrees of freedom.  Each seek lower-fidelity models that are able to accurately predict important statistical moments of the high-fidelity model over longer timescales. As an example, consider a mean-field continuum model derived by coarse-graining a molecular dynamics simulation. Classically, one may pursue homogenization analysis to carefully derive such a continuum model, but such techniques are typically problem specific and can become technical. We illustrate here how GMLS-Nets can be used to extract a conservative continuum PDE model from particle-level simulation data. 

Brownian motion has as its infinitesimal generator the unsteady diffusion equation \cite{karatzas1998brownian}. As a basic example, we will extract a 1D diffusion equation to predict the long-term density of a cloud of particles undergoing pseudo-1D Brownian motion. We consider the periodic domain $\Omega = [0,1]\times [0,0.1]$, and generate a collection of $N_p$ particles with initial position $x_p(t=0)$ drawn from the uniform distribution $U[0,0.5] \times U[0,0.1]$. 

Due to this initialization and domain geometry, the particle density is statistically one dimensional. We estimate the density field $\rho(x,t)$ along the first dimension by constructing a collection $\mathbb{C}$ of $N$ uniform width cells and build a histogram, 
\begin{equation}
\rho(x,t) = \sum_{c \in \mathbb{C}} \sum_{p=1}^{N_p} \mathbbm{1}_{x_p(t) \in c} \mathbbm{1}_{x \in c}.
\end{equation}
The $\mathbbm{1}_{x \in A}$ is the indicator function taking unit value for $x\in A$ and zero otherwise.

We evolve the particle positions $x_p(t)$ under 2D Brownian motion (the density will remain statistically 1D as the particles evolve). In the limit $N_p/N \rightarrow \infty$, the particle density satisfies a diffusion equation, and we can scale the Brownian motion increments to obtain a unit diffusion coefficient in this limit. 

As the ratio $N_p/N$ is finite, there is substantial noise in the extracted density field.  We obtain a low pass filtered density, $\widetilde{\rho}(x,t)$, by convolving $\rho(x,t)$ with a Gaussian kernel of width twice the histogram bin width.

We use the FVM scheme in the same manner as in the previous section.  In particular, we regress a flux that matches the increment ${(\widetilde{\rho}(x,t=10) - \widetilde{\rho}(x,t=12))}/{2\Delta t}$. This window was selected, since the regression at $t=0$ is ineffective as the density approximates a heaviside function.  Such near discontinuities are poorly represented with polynomials and subsequently not expected to train well. Additionally, we train over a time interval of $2 \Delta t$, where in general $k\Delta{t}$ steps can be used to help mollify high-frequency temporal noise. 

To show how the GMLS-Nets' inferred operator can be used to make predictions, we evolve the regressed FVM for one hundred timesteps and compare to the density field obtained from the particle solver.  We apply Dirichlet boundary conditions $\rho(0,t)=\rho(1,t)=1$ and initial conditions matching the histogram $\rho(x,t=0)$.  Again, the FVM by construction is conservative, where it is easily shown for all $t$ that $\int_\Omega \rho dx = N_p$. A time series summarizing the evolution of density in both the particle solver and the regressed continuum model is provided in Fig \ref{brownian}. While this is a basic example, this illustrates the potential of GMLS-nets in constructing continuum-level models from molecular data.  These techniques also could have an impact on data-driven approaches for numerical methods, such as projective integration schemes.

\subsection{Image processing: MNIST benchmark.}
\begin{figure}[h]
\centering
\includegraphics[width=0.6\columnwidth]{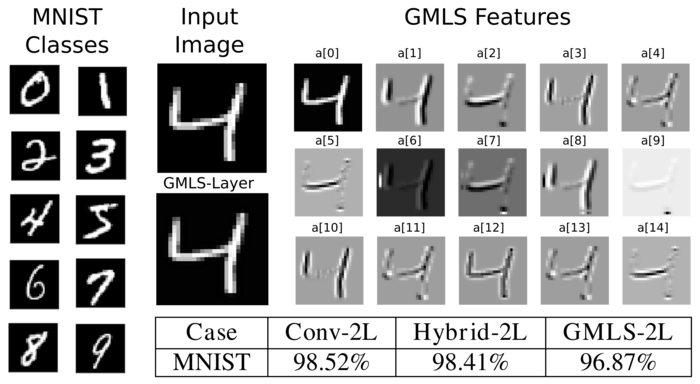}
\caption{MNIST Classification.  GMLS-Layers are substituted for convolution layers in a basic two-layer architecture (Conv2d + ReLu + MaxPool + Conv2d + ReLu + MaxPool + FC). The Conv-2L test are all Conv-Layers, Hybrib-2L has GMLS-Layer followed by a Conv-Layer, and  GMLS-2L uses all GMLS-Layers.
GMLS-Nets used a polynomial basis of monomials.  The filters in GMLS are by design more limited than a general Conv-Layer and correspond here to estimated derivatives of the data set \textit{(top-right)}.  Despite these restrictions, the GMLS-Net still performs reasonably well on this basic classification task (bottom-table).
}
\label{fig:gmls_mnist}
\end{figure}

While image processing is not the primary application area we intend, GMLS-Nets can be used for tasks such as classification.  For the common MNIST benchmark task, we compare use of GMLS-Nets with CNNs in Figure~\ref{fig:gmls_mnist}.   CNNs use kernel size $5$, zero-padding, max-pool reduction $2$, channel sizes $16, 32$, FC as linear map to soft-max prediction of the categories.  The GMLS-Nets use the same architecture with a GMLS using polynomial basis of monomials in $x,y$ up to degree $p_{order} = 4$. 

We find that despite the features extracted by GMLS-Nets being more restricted than a general CNN, there is only a modest decrease in the accuracy for the basic MNIST task.  We do expect larger differences on more sophisticated image tasks.  This basic test illustrates how GMLS-Nets with a polynomial basis extracts features closely associated with taking derivatives of the data field.  We emphasize for other choices of basis for $p^*$ and sampling functionals $\lambda_j$, other features may be extracted.  For polynomials with terms in dictionary order, coefficients are shown in Fig.~\ref{fig:gmls_mnist}.  Notice the clear trends and directional dependence on increases and decreases in the image intensity, indicating $c[1] \sim \partial_x$ and $c[2] \sim \partial_y$.  Given the history of PDE modeling, for many classification and regression tasks arising in the sciences and engineering, we expect such derivative-based features extracted by GMLS-Nets will be useful in these applications.

\subsection{GMLS-Net on unstructured fluid simulation data.}
We consider the application of GMLS-Nets to unstructured data sets representative of scientific machine learning applications. Many hydrodynamic flows can be experimentally characterized using velocimetry measurements.  While velocity fields can be estimated even for complex geometries, in such measurements one often does not have access directly to fields, such as the pressure.  However, integrated quantities of interest, such as drag are fundamental for performing engineering analysis and yet depend upon both the velocity and pressure.  This limits the level of characterization that can be accomplished when using velocimetry data alone.  We construct GMLS-Net architectures that allow for prediction of the drag directly from unstructured fluid velocity data, without any direct measurement of the pressure. 

We illustrate the ideas using flow past a cylinder of radius $L$.  This provides a well-studied canonical problem whose drag is fully characterized experimentally in terms of the Reynolds number, $Re = U L/\nu$. For incompressible flow past a cylinder, one may apply dimensional analysis to relate drag $F_d$ to the Reynolds number via the drag coefficient $C_d$:
\begin{equation}
    \frac{2 F_d}{\rho U_\infty^2 A} =  C_d\left(\frac{U L}{\nu}\right).
\end{equation}
The $U_\infty$ is the free-stream velocity, $A$ is the frontal area of the cylinder, and $C_d:\mathbb{R}\rightarrow\mathbb{R}$.  Such analysis requires in practice engineering judgement to identify relevant dimensionless groups.  After such considerations, this allows one to collapse relevant experimental parameters to ($\rho,U_\infty,A,L,\nu$) onto a single curve. 

\begin{figure}[h]
    \centering
    \includegraphics[width=0.6\linewidth]{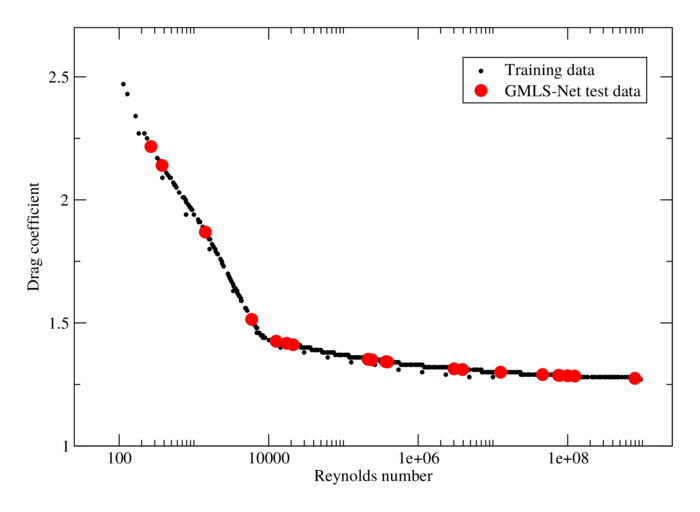}\\
    \includegraphics[width=0.28\linewidth]{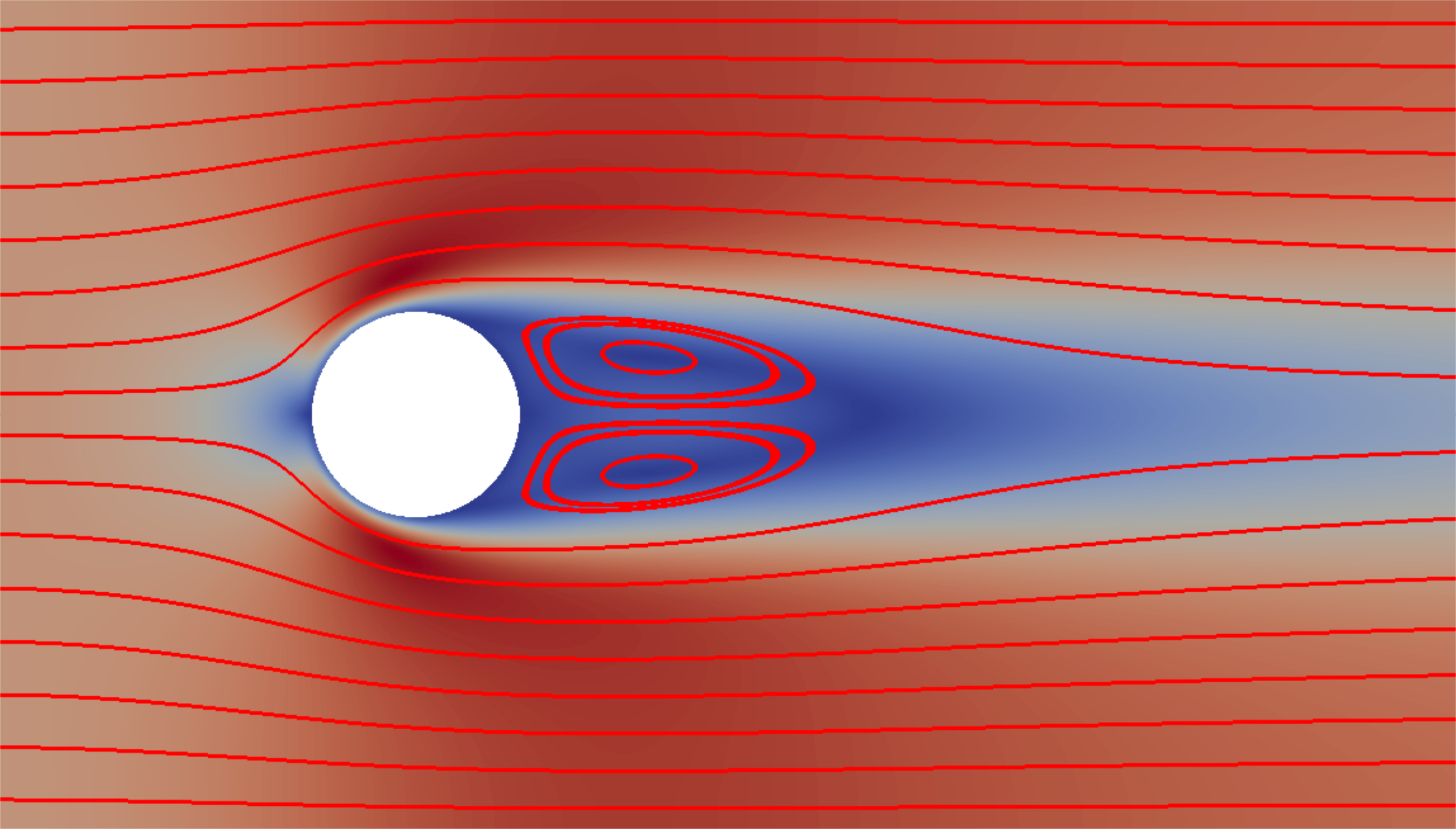}
        \includegraphics[width=0.28\linewidth]{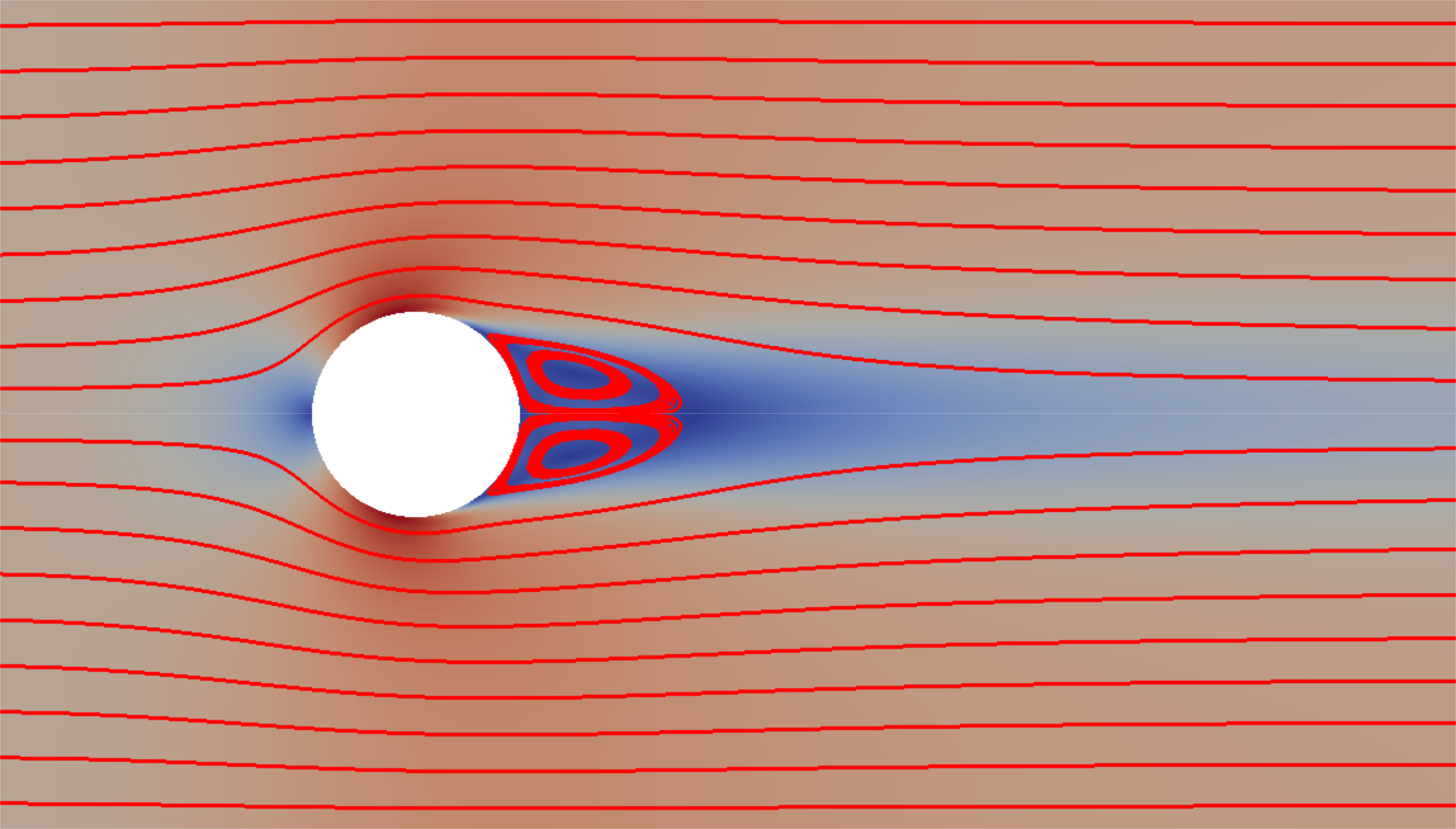}
    \caption{GMLS-Nets are trained on a CFD data set of flow velocity fields.  \textit{Top:} Training set of the drag coefficient plotted as a function of Reynolds number (small black dots). The GMLS-Net predictions for a test set (large red dots). \textit{Bottom:} Flow velocity fields corresponding to the smallest \textit{(left)} and largest \textit{(right)} Reynolds numbers in the test set. }
    \label{fig:drag}
\end{figure}

For the purposes of training a GMLS-Net, we construct a synthetic data set by solving the Reynolds averaged Navier-Stokes (RANS) equations with a steady state finite volume code.  Let $L = \rho = 1$ and consider $U \in \left[0.1,20\right]$ and $\nu \in \left[ 10^{-2},10^{8} \right]$. We consider a $k-\epsilon$ turbulence model with inlet conditions consistent with a $10\%$ turbulence intensity and a mixing length corresponding to the inlet size.  From the solution, we extract the velocity field $\mathbf{u}$ at cell centers to obtain an unstructured point cloud $\mathbb{X}_h$.  We compute $C_d$ directly from the simulations. We then obtain an unstructured data set of $400$ $(\mathbf{u})_i$ features over $\mathbb{X}_h$, with associated labels $C_d$.  We emphasize that although $U_\infty$ and $\nu$ are used to generate the data, they are not included as features, and the Reynolds number is therefore hidden.

We remark that the $k-\epsilon$ model is well known to perform poorly for flows with strong curvature such as recirculation zones. Here, in our proof-of-concept demonstration, we treat the RANS-$k-\epsilon$ solution as ground truth for simplicity, despite its short-comings and acknowledge that a more physical study would consider ensemble averages of LES/DNS data in 3D.  We aim here just to illustrate the potential utility of GMLS-Nets in a scientific setting for processing such unstructured data sets.

As an architecture, we provide two input channels for the two velocity components to three stacked GMLS layers. The first layer acts on the cell centers, and intermediate pooling layers down-sample to random subsets of $\mathbb{X}_h$. We conclude with a linear activation layer to extract the drag coefficient as a single scalar output. We randomly select $80\%$ of the samples for training, and use the remainder as a test set.  We quantify using the root-mean-square (MSE) error which we find to be below $1.5\%$.

The excellent predictive capability demonstrated in Fig. \ref{fig:drag} highlights GMLS-Nets ability to provide an effective means of regressing engineering quantities of interest directly from velocity flow data; the GMLS-Net architecture is able to identify a latent low-dimensional parameter space which is typically found by hand using dimensional analysis.  This similarity relationship across the Reynolds numbers is identified, despite the fact that it does not have direct access to the viscosity parameter.  These initial results indicate some of the potential of GMLS-Nets in processing unstructured data sets for scientific machine learning applications.

\section{Conclusions}
We have introduced GMLS-Nets for processing scattered data sets leveraging the framework of GMLS.  GMLS-Nets allow for generalizing convolutional networks to scattered data, while still benefiting from underlying translational invariances and weight sharing.  The GMLS-layers provide feature extractors that are natural particularly for regressing differential operators, developing dynamical models, and predicting 
quantities of interest associated with physical systems.  GMLS-Nets were demonstrated to be capable of obtaining dynamical models for long-time integration beyond the limits of traditional CFL conditions, for making predictions of density evolution of molecular systems, and for predicting directly from flow data quantities of interest in fluid mechanics.  These initial results indicate some promising capabilities of GMLS-Nets for use in data-driven modeling in scientific machine learning applications.  

\section*{} 
\small

\printbibliography 

\appendix 

\clearpage 
\newpage

\section{Derivation of Gradients of the Operator  $\tau_{x_i}[u]$.}

\subsection{Parameters of the operator $\tilde{\tau}$.}

We give here some details on the derivation of the gradients for the learnable GMLS operator $\tau[u]$ and intermediate steps.  This can be used in implementations for back-propagation and other applications.

GMLS works by mapping data to a local polynomial fit in region $\Omega_i$ around $x_i$ with $p^*(x) \approx u(x)$ for $x \in \Omega_i$.  To find the optimal fitting polynomial $p^*(x) \in \mathcal{V}$ to the function $u(x)$, we can consider the case with $\lambda_j(x) = \delta(x - x_j)$ and weight function $w_{ij} = w(x_i - x_j)$.  In a region around a reference point $x^*$ the optimization problem can be expressed parameterically in terms of coefficients $\mathbf{a}$ as
$$
\mathbf{a}^*(x_i) =
\arg \min_{\mathbf{a} \in \mathbb{R}^m} 
\sum_j 
\left(
u_j - \mathbf{p}(x_j)^T \mathbf{a}
\right)^2 w_{ij}.
$$
We write for short $\mathbf{p}(x_j) = \mathbf{p}(x_j,x_i)$, where the basis elements in fact do depend on $x_i$.  Typically, for polynomials we just use $\mathbf{p}(x_j,x_i) = \mathbf{p}(x_j - x_i)$.  This is important in the case we want to take derivatives in the input values $x_i$ of the expressions.

We can compute the derivative in $a_\ell$ to obtain 
$$
\frac{\partial J}{\partial a_\ell}(x_i) = 0.
$$
This implies
$$
\left[\sum_j \mathbf{p}(x_j) w_{ij} \mathbf{p}(x_j)^T \right] \mathbf{a} = \sum_j  w_{ij} \mathbf{p}(x_j) u_j.
$$
Let
$$
M = \left[\sum_j \mathbf{p}(x_j) w_{ij} \mathbf{p}(x_j)^T \right], \hspace{0.25cm} \mathbf{r} = \sum_j  w_{ij} \mathbf{p}(x_j) u_j,
$$
then we can rewrite the coefficients as the solution of the linear system 
$$
M\mathbf{a}^*(x_i) = \mathbf{r}.
$$
This is sometimes written more explicitly for analysis and computations as 
$$
\mathbf{a}^*(x_i) = M^{-1}\mathbf{r}.
$$
We can represent a general linear operator $\tilde{\tau}(x_i)$ using the $\mathbf{a}^*$ representation as 
$$
\tilde{\tau}(x_i) = \mathbf{q}(x_i)^T \mathbf{a}^*(x_i)
$$
Typically, the weights will not be spatially dependent $\mathbf{q}(x_i) = \mathbf{q}_0$. Throughout, we shall denote this simply as $\mathbf{q}$ and assume there is no spatial dependence, unless otherwise indicated.

\subsection{Derivatives of $\tilde{\tau}$ in ${x_i}$, ${\mb{a}(x_i)}$, and $\mb{q}$.}

The derivative in $x_i$ is given by
$$\frac{\partial}{\partial x_i} \mathbf{a}^*(x_i) = \frac{\partial M^{-1}}{\partial x_i} \mathbf{r} + M^{-1}\frac{\partial \mathbf{r}}{\partial x_i}$$
In the notation, we denote $\mathbf{p}(x_j) = \mathbf{p}(x_j,x_i)$, where the basis elements in fact can depend on the particular $x_i$.  These terms can be expressed as 
$$\frac{\partial M^{-1}}{\partial x_i} = -M^{-2}\frac{\partial M^{-1}}{\partial x_i},$$
where
\begin{eqnarray}
\nonumber
\frac{\partial M}{\partial x_i} &=& 
\sum_j \left[\left(\frac{\partial }{\partial x_i}\mathbf{p}(x_j,x_i)\right)\mathbf{p}(x_j,x_i)^T w_{ij} 
\right. \\
\nonumber
&+& \mathbf{p}(x_j,x_i)\left(\frac{\partial }{\partial x_i}\mathbf{p}(x_j,x_i)\right)^T w_{ij} \\
&+&
\nonumber
\left.
\mathbf{p}(x_j,x_i)\mathbf{p}(x_j,x_i)^T \frac{\partial w_{ij}}{\partial x_i}\right].
\end{eqnarray}
The derivatives in $\mb{r}$ are given by
$$\frac{\partial \mathbf{r}}{\partial x_i} = 
\sum_j\left[
\left(\frac{\partial }{\partial x_i}\mathbf{p}(x_j)\right)u_j w_{ij}
+
\mathbf{p}(x_j)u_j \frac{\partial w_{ij}}{\partial x_i}
\right]
.$$
The full derivative of the linear operator $\tilde{\tau}$ can be expressed as 
$$
\frac{\partial}{\partial x_i} \tilde{\tau}(x_i) 
= \left(\frac{\partial}{\partial x_i} \mathbf{q}(x_i)^T\right) \mathbf{a}^*(x_i) 
+ \mathbf{q}(x_i)^T \left(\frac{\partial}{\partial x_i} \mathbf{a}^*(x_i)\right).
$$
In the constant case $\mathbf{q}(x_i) = \mathbf{q}_0$, the derivative of $\tilde{\tau}$ simplifies to
$$
\frac{\partial}{\partial x_i} \tilde{\tau}(x_i) 
= \mathbf{q}_0^T \left(\frac{\partial}{\partial x_i} \mathbf{a}^*(x_i)\right).
$$

The derivatives of the other terms follow more readily.  For derivative of the linear operator $\tilde{\tau}$ in the coefficients $\mb{a}(x_i)$, we have
$$
\frac{\partial}{\partial \mathbf{a}(x_i)} \tilde{\tau}(x_i) 
= \mathbf{q}(x_i).
$$
For derivatives of the linear operator $\tilde{\tau}$ in the mapping coefficient $\mb{q}$ values, we have
$$
\frac{\partial}{\partial \mathbf{q}(x_i)} \tilde{\tau}(x_i) 
= \mathbf{a}(x_i).
$$

In the case of nonlinear operators $\tilde{\tau} = \mb{q}(\mathbf{a}(x_i))$ there are further dependencies beyond just $x_i$ and $\mb{a}(x_i)$, and less explicit expressions.  For example, when using MLP's there may be hierarchy of trainable weights $\mb{w}$.  The derivatives of the non-linear operator can be expressed as
$$
\frac{\partial}{\partial \mb{w}} \tilde{\tau}(x_i) 
= \frac{\partial \mb{q}}{\partial \mb{w}}(\mathbf{a}(x_i)).
$$
Here, one relies on back-propagation algorithms for evaluation of $\frac{\partial \mb{q}}{\partial \mb{w}}$.  Similarly, given the generality of $\mb{q}(\mb{a})$, 
for derivatives in $\mb{a}$ and $x_i$, one can use back-propagation methods on $\mb{q}$ and the chain-rule with the expressions derived during the linear case for $\mb{a}$ and $x_i$ dependencies.

\end{document}